\RequirePackage[2020-02-02]{latexrelease}
\documentclass[twocolumn, switch]{article} 

\usepackage{arxiv}

\usepackage{amsmath, amsthm, amssymb, amsfonts}

\usepackage{subfig, graphicx}

\usepackage{algorithm}
\usepackage{algpseudocode}

\usepackage[numbers,square]{natbib}
\usepackage[utf8]{inputenc}	
\usepackage[T1]{fontenc}	
\usepackage{xcolor}		
\usepackage[colorlinks = true,
            linkcolor = purple,
            urlcolor  = blue,
            citecolor = cyan,
            anchorcolor = black]{hyperref}	
\usepackage{booktabs} 		
\usepackage{nicefrac}		
\usepackage{microtype}		
\usepackage{lineno}		
\usepackage{float}			

\usepackage{lipsum}		

\usepackage{newfloat}
\DeclareFloatingEnvironment[name={Supplementary Figure}]{suppfigure}
\usepackage{sidecap}
\sidecaptionvpos{figure}{c}

\usepackage{titlesec}
\titlespacing\section{0pt}{12pt plus 3pt minus 3pt}{1pt plus 1pt minus 1pt}
\titlespacing\subsection{0pt}{10pt plus 3pt minus 3pt}{1pt plus 1pt minus 1pt}
\titlespacing\subsubsection{0pt}{8pt plus 3pt minus 3pt}{1pt plus 1pt minus 1pt}

\usepackage{tikz,xcolor,hyperref}

\definecolor{lime}{HTML}{A6CE39}
\DeclareRobustCommand{\orcidicon}{
	\begin{tikzpicture}
	\draw[lime, fill=lime] (0,0) 
	circle [radius=0.16] 
	node[white] {{\fontfamily{qag}\selectfont \tiny ID}};
	\draw[white, fill=white] (-0.0625,0.095) 
	circle [radius=0.007];
	\end{tikzpicture}
	\hspace{-2mm}
}
\foreach \x in {A, ..., Z}{\expandafter\xdef\csname orcid\x\endcsname{\noexpand\href{https://orcid.org/\csname orcidauthor\x\endcsname}
			{\noexpand\orcidicon}}
}

\title{\textit{Can we Agree?} On the Rashōmon Effect and the Reliability of Post-Hoc Explainable AI}

\usepackage{xwatermark}
\newwatermark[firstpage,color=gray!90,angle=0,scale=0.28, xpos=0in,ypos=-5in]{*correspondence: \texttt{marion.noulhiane@cea.fr}}

\usepackage{authblk}

\author[1,2,3,4]{Clément Poiret\orcidA{}}
\author[3]{Antoine Grigis\orcidC{}}
\author[2]{Justin Thomas}
\author[1,3,4\thanks{\tt{marion.noulhiane@cea.fr}}]{Marion Noulhiane\orcidD{}}

\affil[1]{UNIACT, NeuroSpin, CEA Paris-Saclay, Frederic Joliot Institute, Gif-sur-Yvette, France}
\affil[2]{Department of Research and Development, Caminov, Paris, France}
\affil[3]{NeuroSpin, CEA Paris-Saclay, Frederic Joliot Institute, Gif-sur-Yvette, France}
\affil[4]{InDEV, NeuroDiderot, Université Paris Cité, Inserm, Paris, France}

\begin{document}

\twocolumn[ 
	\begin{@twocolumnfalse} 
				  
		\maketitle
				
		\begin{abstract}
    		The Rashōmon effect poses challenges for deriving reliable knowledge from machine learning models. This study examined the influence of sample size on explanations from models in a Rashōmon set using SHAP. Experiments on 5 public datasets showed that explanations gradually converged as the sample size increased. Explanations from <128 samples exhibited high variability, limiting reliable knowledge extraction. However, agreement between models improved with more data, allowing for consensus. Bagging ensembles often had higher agreement. The results provide guidance on sufficient data to trust explanations. Variability at low samples suggests that conclusions may be unreliable without validation. Further work is needed with more model types, data domains, and explanation methods. Testing convergence in neural networks and with model-specific explanation methods would be impactful. The approaches explored here point towards principled techniques for eliciting knowledge from ambiguous models.
		\end{abstract}
		\vspace{0.35cm}
				
	\end{@twocolumnfalse} 
] 



\section{Introduction}

In recent years, the widespread integration of AI-powered systems in various domains has highlighted the need for increased transparency and trust in these complex models. Explainable AI (XAI) methodologies have emerged as a means to address this need by providing insights into model training and clarifying predictions through post hoc explanations. As these systems become more integrated into our daily lives and are employed in critical domains such as healthcare (e.g. \cite{grzybowskiApprovalCertificationOphthalmic2023,pandeyTransformationalRoleGPU2022}) and cybersecurity (e.g. \cite{dixitDeepLearningAlgorithms2021}), it is crucial to understand the basis of their decision-making processes. By unraveling the inner workings of these complex models, we can ascertain whether their conclusions are based on genuine features or merely rely on spurious correlations and shortcuts. In addition, ethical considerations play a role in the adoption and deployment of these AI systems. As they become increasingly sophisticated, it is essential to ensure that their decision-making processes are in line with ethical guidelines and do not follow biases or discriminate against certain individuals or communities. Explainable AI methodologies, such as the post hoc explanations provided by XAI techniques, offer a means of scrutinizing and mitigate potential biases and ethical concerns. By shedding light on the decision-making process and revealing the factors that contribute to predictions, XAI not only provides accountability, but also empowers stakeholders to address and rectify any biases or ethical issues that may arise. Beyond ethical concerns, if these models outperform human experts on key metrics, illuminating the black box could lead to new findings that might enhance expertise, expand knowledge, or spark new research ideas.

While there is an active research field dedicated to building explainable models out-of-the-box, those models are not yet widely available or as effective as state-of-the-art black-box models in some domains. Therefore, this current work will focus solely on post-hoc explanations: methods that will link a model's prediction to contextual information. Unfortunately, multiple models can perform very well on a given task, but there is no guarantee on their use of similar features to construct their prediction. Multiple models can base their predictions on mutiple set of different features, an effect named ``the Rashōmon Effect'' 

\subsection{Explainable AI: a Practical Overview}

\begin{figure}[!htb]
	\centering
	\includegraphics[width=\columnwidth]{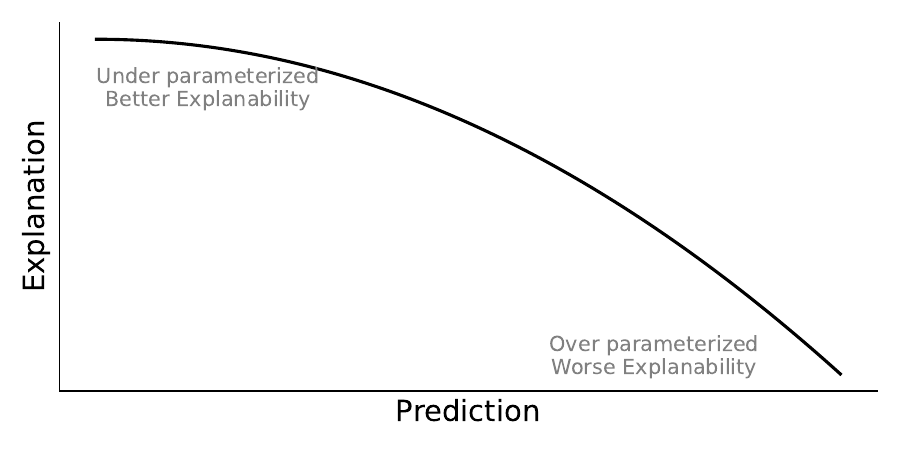}
	\caption{Usual Perception of the Prediction-Explanation trade-off. The prediction-explanation trade-off illustrates the difficulty behind XAI. The more complex a model is, the more predictive power it has, but its use of highly complex structures render its interpretability nearly intractable.}
	\label{fig:p-e_fallacy}
\end{figure}

Although interpretability and explainability may be used interchangeably, interpretability refers to the ability to understand the internal workings and mechanisms of a model, whereas explainability, on which this work will focus, refers to the capacity to provide explanations for the model's predictions or decisions in a way that can be understood by humans. However, as Del Giudice et al., (2022) previously described \cite{del_giudice_prediction-explanation_2021}, there exists a Prediction-Explanation Fallacy. It arises when one employs prediction-optimized models for explanatory purposes, without taking into account the delicate balance between explanation and prediction. On the one hand, of the spectrum lie interpretable models, which are either unrealistically simplistic or complex to deploy in real-world situations. On the other hand, the models employ excessively complex structures that are almost impossible to interpret (\autoref{fig:p-e_fallacy}). Thus, if explainability can come from models with built-in explanation mechanisms (e.g. ref), the easiest approach comes from post hoc explanation methods, as they allow the explanation of any given model, for any given task. Finally, the objectives of explainability are (i) to provide information on the training or generalization of a specific model, or (ii) to clarify its predictions by explaining them in terms of the input of the model \cite{rasExplainableDeepLearning2022}. By addressing the question of \textit{``why did this model make this particular prediction?''}, XAI generates a set of features that highlight distinct patterns present in a model's internal representation of a phenomenon.

Post-hoc explainability encompasses two main approaches: model-specific and model-agnostic methods. Model-specific methods are tailored to a specific type of model and provide insight into its internal workings. For example, attention maps (e.g. ref) highlight the important regions in an image that contribute to the model's prediction, while GradCAM (e.g. \cite{selvarajuGradCAMVisualExplanations2020}) visualizes the regions that are crucial for a specific class. On the other hand, model-agnostic methods are applicable to any model and focus on explaining predictions without accessing internal information such as weights or architecture \cite{carvalho_machine_2019}. SHAP (for SHapley Additive exPlanations) \cite{NIPS2017_7062} - the most popular method of this kind - uses Shapley values to attribute the importance of each input feature in making the final prediction. For example, in an image classification task, SHAP can identify the pixels that contribute the most to the predicted class. Shapley values are a way to fairly distribute the ``credit'' for a particular outcome among the different factors that contributed to it. In the context of Machine and Deep Learning, ``credit'' refers to the importance of each input feature (e.g. pixels in an image) in making the final prediction. SHAP computes the Shapley value for each feature by considering all possible combinations of features and measuring the change in prediction.

In addition to the increase in trust and transparency, explaining a model gives the possibility of detecting inconsistencies in its modeling quality. Some models may demonstrate equivalent predictive accuracy, but their biases may be distinct, implying the ``Rashōmon effect'', --- i.e. conflicting interpretations of the underlying phenomenon \cite{breiman_statistical_2001}.

\subsection{On the Rashōmon Effect}

The tension between prediction and explanation is correlated to the Rashōmon Effect. According to Anderson (2016), the Rashōmon Effect ``is the naming of an epistemological framework—or ways of thinking, knowing, and remembering—required for understanding the complex and ambiguous on both the small and large scale'' \cite{anderson_rashomon_2016}. 
It originates from a Japanese film of the same name, by Akira Kurosawa, which portrays a crime from four different perspectives, each with a different interpretation of the event caused by the subjectivity of human perception and memory. Thus, Breiman (2001) imported the concept into statistics and Machine Learning, where the concept refers to the phenomenon in which different near-optimal models trained on the same task may actually base their prediction on different sets of features \cite{breiman_statistical_2001}. Different models that perform similarly may have different underlying structures and assumptions.

\begin{figure}[!htb]
	\centering
	\includegraphics[width=\columnwidth]{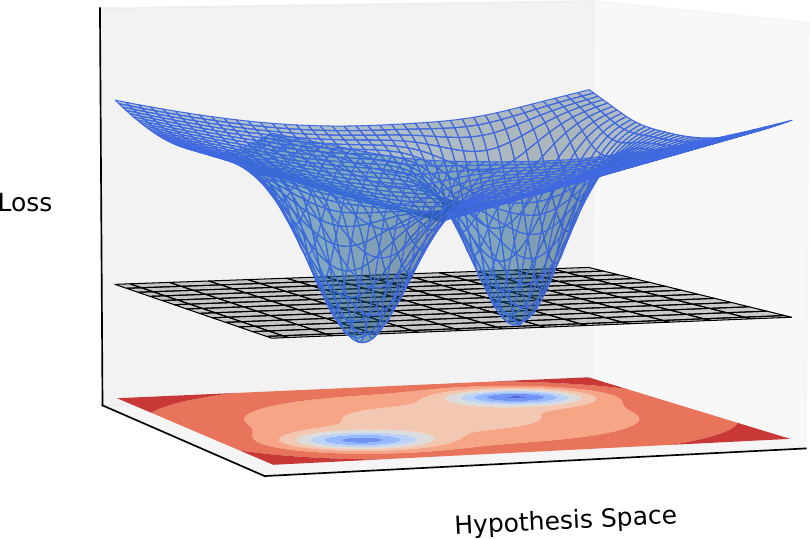}
	\caption{A simplified example of a Rashōmon set in a two-dimensional hypothesis space. The loss of each model in the space is in blue, while the black plan represents the value of the Rashōmon parameter, $\epsilon$. As a result, the Rashōmon set is illustrated in blue in the projection at the bottom of the figure.}
	\label{fig:Rashōmon-set}
\end{figure}

Those models fall into what we call a Rashōmon set: a set of models showing the same predictive power, i.e., models whose training loss is below a specific threshold. Given a loss function $L$ and a model class $F$, the Rashōmon set $R$ can be written as \cite{rudin_interpretable_2021}:

$$R(F, f^*, \epsilon) =  \{ f \in F \text{ such that } L(f) \leq L(f^*) + \epsilon \} $$

where $f^*$ is an optimal model, $\epsilon$ is a threshold named the Rashōmon parameter. While multiple models can be found at $\epsilon = 0$, we usually use and $\epsilon$-level set where $\epsilon > 0$ \cite{marx_predictive_2020} because a low $\epsilon$ could result in an empty set. It has to be noted that computing the Rashōmon set is NP-hard and requires brute-force methods by sampling the hypothesis space \cite{rudin_interpretable_2021}.

Let $H$ be the hypothesis space that contains all possible models $M$, and let $f_1$ and $f_2$ be two near-optimal models in $R(F, f^*, \epsilon)$ of $H$. The Rashōmon Effect refers to the phenomenon where $f_1$ and $f_2$ produce similar predictions on the same input $x$, despite showing different feature importances. Formally, with $w_1$ and $w_2$ be the feature importances of $f_1$ and $f_2$ respectively, let $r_1$ and $r_2$ be the ranks of each feature. Intuitively, we can say that we observe a Rashōmon effect when:

$$r_1 \neq r_2$$

The Rashōmon Effect, in this case, can be problematic because it can lead to different interpretations of the same dataset and make it difficult to identify the most important features for a given task, or even to derive reliable knowledge from this interpretation.

\subsection{Finding a Consensus}

As formulated by Teney et al. (2022), predicting is not understanding \cite{teney_predicting_2022}. The presence of the Rashōmon effect can be seen as a manifestation of underspecification, where a model fails to capture all the underlying patterns in the data accurately. To address this issue, Teney et al. propose training multiple models that are compatible with the data. Del Giudice (2022) suggests several ways to mitigate this problem, notably: (i) seeking consensus among models with different assumptions or biases, and (ii) noting that the severity of this effect tends to decrease when working with larger datasets \cite{del_giudice_prediction-explanation_2021}.

Thus, we hypothesize that: 

\begin{enumerate}
    \item if computing the Rashōmon set without brute force is challenging, it is still possible to identify models that belong to the set,
    \item the robustness against small input perturbations increases as the sample size grows,
    \item the agreement among explanations from diverse models in the Rashōmon set converges, allowing for a consensus when the sample size is sufficiently large.
\end{enumerate}

To further enhance the exploration of the Rashōmon effect, we included a bootstrap aggregation (bagging) strategy, a common method that enhances overall performance by alleviating fluctuations in predictions from multiple models, as we can suppose that they do so also by minimizing the variability of feature importances. To investigate the said hypothesis, the work has been organized into three main contributions on five distinct public datasets.

\begin{enumerate}
    \item we proposed a methodology to assess an intra-model agreement, in which the explainability is disturbed by a varying input dataset through cross-validation,
    \item we proposed a methodology to assess an inter-model agreement, and its convergence toward a consensus,
    \item we analyzed the behavior of a bagging strategy to assess its behavior with respect to intra- and inter-model agreement.
\end{enumerate}

\begin{figure*}[!htb]
	\centering
	\includegraphics[width=\textwidth]{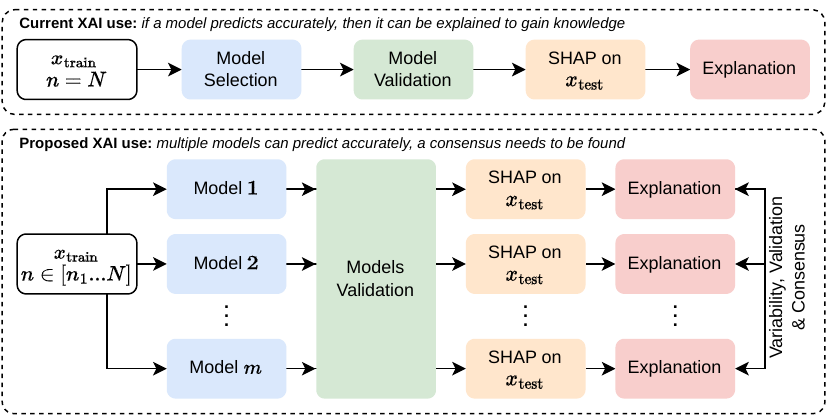}
	\caption{Proposed framework for enhancing explainability through model validation and consensus finding. It illustrates the proposed framework to improve the reliability of explanations from machine learning models. In contrast to the typical use of XAI methods where a single predictive model is explained, the proposed approach first validates multiple high-performing models on a dataset. SHAP explanations are generated for each model in the validation stage. The resulting explanations are then analyzed to find consensus and reduce variability, leading to more trustworthy explanations.}
	\label{fig:framework}
\end{figure*}

The intent of this paper is not to benchmark explanation methods, but instead presents a novel approach to evaluating the robustness of explanations, notably with respect to sample size. The general goal of the present work is to propose a new framework for computing explanations of machine learning models (\autoref{fig:framework}). Its aim is to emphasize the fact that explaining a model is not enough, but we need to assess the convergence and robustness of feature importances.

\section{Methodology}
\label{sec:methodology}

The method (\autoref{fig:framework}) can be divided into four parts: (i) a pre-selection of models based on their performances on the training set (\autoref{table:models}) followed by their hyperparameter tuning to ensure we ended up in a Rashōmon set; (ii) a 10-fold cross-validation on subsets of the training set with varying size of each model, to ensure that the models are generalizing well; (iii) a bagging strategy on all the selected models, which is a technique that combines the predictions of multiple models to improve the overall accuracy; and (iv) an inference on a holdout test set followed by an explanation using the SHAP methodology. All training, inference, and explanation steps are summarized in algorithm \autoref{alg:method}. As sample size might have an impact on hyperparameters, we conducted a simple hyperparameter tuning using a random search strategy for each instance of a model, to ensure that the best hyperparameters are chosen for each model.

\renewcommand{\algorithmicrequire}{\textbf{Input:}}
\renewcommand{\algorithmicensure}{\textbf{Output:}}
\begin{algorithm}
\caption{Model Selection and Explanations}\label{alg:method}
    \begin{algorithmic}[1]
        \Require{$x_{train}$, $x_{test}$ (training and test sets)}
        \Ensure{$sim_{intra}, sim_{inter}$ (similarities between explanations)}
        \Procedure{Select}{$M$}
            \State \texttt{r = []} \Comment{Array of Kappa scores}
            
            \For{\texttt{$f \in M$}}
                \State \texttt{Train($f,x_{train}$)}
                \State $\hat{y} = f(x_{test})$
                \State $\mathcal{K} = \texttt{Kappa}(\hat{y},y)$
                \State \texttt{r.append}($\mathcal{K}$)
            \EndFor

            \State $top_3 = \texttt{Sort(Rank(r))}[::-1][3]$
        \EndProcedure

        \Procedure{Explain}{$top_3$}

        \State $S = \{2^4, 2^5, ..., N\}$ \Comment{Varying sample sizes}

        \State $\Phi \texttt{= []}$ \Comment{Array of explanations}
        \For{\texttt{$s \in S$}}
            \State $x = x_{train}[:s]$
            
            \For{\texttt{$f \in top_3$}} \Comment{Repeated as 10-Fold CV}
                \State $\texttt{Train($f, x$)}$
                \State $\hat{y} = f(x_{test})$
                \State $\phi^{SHAP}_f = \texttt{Explain($\hat{y}$)}$
                \State \texttt{$\Phi$.append($\phi^{SHAP}_f$)}
            \EndFor
        \EndFor

        \State $sim_{intra} = similarity(\Phi|s)$ \Comment{For each model, computes the similarity between cross-validation folds}
        \State $sim_{inter} = similarity(\Phi|f)$ \Comment{For each sample size, computes the similarity between all the models}

        \EndProcedure
    
    \end{algorithmic}
\end{algorithm}

\subsection{Datasets}

To test our hypothesis under various realistic conditions, we conducted our experiments on five datasets (\autoref{table:datasets}, of various sizes, dimensionality, and with or without class imbalance. Most of them come from the OpenXAI framework, an open initiative for the robust and repeatable evaluation of XAI methodologies \cite{agarwalOpenXAITransparentEvaluation2023}.

\begin{table*}[!htb]
	\centering
	\begin{tabular}{lllll}
		\hline
		Dataset                   & Short name & N     & N Features    & Balanced  \\  \hline
		Framingham Heart study \cite{ashishbhardwajFraminghamHeartStudy}    & Framingham & 4240  & 16            & No        \\
            German Credit \cite{hofmannStatlogGermanCredit1994}            & German     & 1000  & 20            & No        \\
            Pima-Indians Diabetes \cite{smithUsingADAPLearning1988}     & Diabetes   & 768   & 9             & No        \\
            COMPAS \cite{jordanEffectRaceEthnicity2015}                    & Compas     & 18876 & 7             & No        \\
            Student \cite{agarwalOpenXAITransparentEvaluation2023}                   & Student    & 100   & 30            & Yes       \\  \hline
	\end{tabular}
	\caption{Datasets used to study the Rashōmon effect.}
	\label{table:datasets}
\end{table*}

\subsection{Exploring the Rashōmon Set}

The first step of this methodology is to select the models that we will explain. As the exploration of the Rashōmon set is NP-hard, its space has been divided into categories of models $M$ (see \autoref{table:models}). Using the PyCaret library \cite{PyCaret}, each of these models has been benchmarked with multiple configurations using 10-fold cross-validation, and the three configurations with the highest Cohen's Kappa coefficient have been selected. The Cohen's Kappa coefficient \cite{cohenCoefficientAgreementNominal1960} is a statistical measure of the interrater agreement taking into account the possibility of agreement occurring by chance. This is particularly important in cases where the classes are imbalanced. In addition to these top 3 models, a bagging strategy has been implemented by combining the predictions of the aforementioned models to further improve the overall comparison.

\begin{table}[!htb]
	\centering
        \resizebox{\columnwidth}{!}{%
    	\begin{tabular}{lllll}
    		\hline
    		Model                           & Acronym  & Linearity    \\  \hline
    		Logistic Regression             & lr       & Linear       \\
                Linear Discriminant analysis    & lda      & Linear       \\
                Ridge                           & ridge    & Linear       \\
                Gaussian Naive Bayes            & nb       & Linear       \\
                Singular Vector Machine         & svm      & Linear       \\
                Singular Vector Machine (RBF)   & rbfsvm   & Non-linear   \\
                Decision Tree                   & dt       & Non-linear   \\
                Quadratic Discriminant analysis & qda      & Non-linear   \\
                Random Forest                   & rf       & Non-linear   \\
                AdaBoost                        & ada      & Non-linear   \\
                Extra Trees                     & et       & Non-linear   \\
                Gradient Boosting               & gbc      & Non-linear   \\
    		K-Nearest Neighbors             & knn      & Non-linear   \\
    		Gaussian Process                & gp       & Non-linear   \\
    		CatBoost                        & catboost & Non-linear   \\
    		LightGBM                        & lightgbm & Non-linear   \\
    		XGBoost                         & xgboost  & Non-linear   \\  \hline
    	\end{tabular}}
        \vspace{2mm}
	\caption{List of all models included in our experiments. For each dataset, all models are trained and tuned, and the best ones are kept for further analysis.}
	\label{table:models}
\end{table}

\subsection{Shapley Additive Explanations (SHAP)}

After model selection and training, the core of our hypothesis relies on feature importances that have to be comparable across all models. As all datasets are already given by the original authors as separate train and a test sets, each model inferred on the latter. Each model state - i.e. each cross-validation step at each sample size - went through the SHAP pipeline \cite{lundbergUnifiedApproachInterpreting2017}. We chose the SHAP method for its ease of use, widespread adoption, and its axiomatic superiority to its counterpart. Shapley-based explanation methods provide a way to assign each feature an importance in a model-agnostic way \cite{lundbergUnifiedApproachInterpreting2017}, allowing the comparison between very different models that could have been previously selected. It has to be noted that, while SHAP is a methodological choice, the method we introduce could have been used with any other explanation framework. Following the definition of \cite{johnsenInferringFeatureImportance2023}, let $\phi^{SHAP}_{f_i}$ be the SHAP values obtained from a machine learning model $f_i$, for a set of $n$ instances and their labels $y$ in a dataset $X$ of $K$ features. In simpler terms, SHAP values are used to explain predictions made by a model at an individual level. Transitioning from the individual level to the population level can be achieved by taking the mean of the absolute values of the vector $\phi^{SHAP}_{f_i}$ across all instances:

$$\frac{1}{N}\sum_{n=1}^{N}|{\phi^{SHAP}_{f_i}}_n|$$

\subsection{Evaluating the similarity between multiple explanations}

As our main goals were to compare the SHAP values given by (i) the same models with variations in the training set induced by cross-validation, and (ii) different near-optimal models on the same training set, we defined two metrics. First, since not all features may be of interest, we used a $top_j$ similarity as introduced in \cite{messalasModelAgnosticInterpretabilityShapley2019}. Then, we used a weighted cosine similarity, which allows us to compare the similarity between multiple vectors while taking into account that features that are not important are possibly randomly ranked by a given model $f_i$.

\subsubsection{Metrics}

\textbf{$top_j$ Similarity}

The $top_j$ similarity metric is a useful tool for comparing the similarity between two sets of features selected by different models. This metric is based on the idea that not all features may be of interest, and therefore, we can focus on the top $j$ features selected by each model. When comparing the overlap between these top $j$ features, we can obtain a measure of similarity between the two models. This metric is particularly useful when comparing models with different feature selection methods or when only a subset of features is relevant to the problem at hand. The $top_j$ similarity metric is a simple, yet effective way to compare the interpretability of different machine learning models and can provide valuable insight into the behavior of these models. To begin with, let us define the ranking function $R$ that maps the SHAP values to the ranked features as follows:

\begin{equation}
	R(\phi^{SHAP}_{f_i} = \text{argsort}(|\phi^{SHAP}_{f_i}|)[::-1][:j]
\end{equation}

where $\text{argsort}(a)$ returns the indices that would sort the array $a$ in ascending order, and the indexing operation $[::-1]$ reverses the order of the sorted indices to obtain the descending order. The final indexing operation $[:j]$ selects the top $j$ indices based on this ranking. 

Using this ranking function, we can obtain the sets of top $j$ features selected by models $f_1$ and $f_2$ for the $n$-th instance as $S^{f_1}_n = R(\phi^{{SHAP}_{f_1}}_n)$ and $S^{f_2}_n = R(\phi^{{SHAP}_{f_2}}_n)$. The $top_j$ similarity metric is the ratio of the average number of common features between the two models and $j$, where $j$ is the number of features selected. Thus, the $top_j$ similarity metric can be expressed as:

$$top_j \text{ similarity} = \frac{1}{j} \sum_{i=1}^{j} \frac{\left| S^{f_1}_i \cap S^{f_2}_i \right|}{\left| S^{f_2}_i \right|}$$

Here, the numerator represents the number of common features between the two models, and the denominator represents the number of features selected by the second model. As we compare $M$ models, the final metric $top^M_j$ computes the mean $top_j$ similarity between all possible combinations of two different models using their respective SHAP values, such as:

\begin{equation}
	top^M_j \text{ similarity} = \frac{1}{\binom{M}{2}} \sum_{a,b \in \binom{[1, M]}{2}} top_j\text{ similarity}(a,b)
\end{equation}

where $\binom{M}{2}$ is the number of possible combinations of two different models, and $\binom{[1, M]}{2}$ is the set of all such combinations.

\textbf{Weighted Cosine Similarity}

We introduced the weighted cosine similarity as a useful metric to compare similarity between multiple vectors while considering that features that are not important are possibly randomly ranked by a given model $f_i$. First, with $K$ features, we compute the mean absolute SHAP value for each feature $k$ across all samples as:

$$mas_k = \frac{1}{K} \sum_{k=1}^{K} |\phi^{SHAP}_{f_{i, k}}|$$

Then, we compute the weighted SHAP values for each feature $k$ as:

$$w_{f_{i, k}} = |\phi^{SHAP}_{f_{i, k}}| \cdot mas_k$$

where $k \in \{1, K\}$. Let us define the weighted cosine similarity $wcossim$ between models $f_1$ and $f_2$ such as:

$$\text{wcossim}(w_{f_1}, w_{f_2}) = \frac{w_{f_1} \cdot w_{f_2}}{\|w_{f_1}\|_2 \|w_{f_2}\|_2}$$

where $\cdot$ denotes the dot product and $\|\cdot\|_2$ denotes the Euclidean norm. Similarly to $top_j$ similarity, we can finally define $wcossim^m$ as the weighted cosine similarity between all combinations of two different models:

\begin{equation}
	\text{wcossim}^M = \frac{1}{\binom{m}{2}} \sum_{a,b \in \binom{[1, M]}{2}} \text{wcossim}(a,b)
\end{equation}

We used the weighted cosine similarity to compare the similarity of models' SHAP values while accounting for the randomness of unimportant feature ranking.

\subsection{Assessing the convergence towards a near-optimal consensus}


The mean of the absolute SHAP values for the largest sample size $N$ available from all models $f_i$ can be expressed as:

\begin{equation}
	\bar{\phi}^{SHAP}_{consensus} = \frac{1}{MN}\sum_{i=1}^{M}\sum_{j=1}^{N} |\phi^{SHAP}_{f_i,j}|
\end{equation}

where $\bar{\phi}^{SHAP}_{consensus}$ represents the mean absolute SHAP values for the consensus of all models, $M$ is the total number of models, $N$ represents the largest sample size available across all models, and $|\phi^{SHAP}_{f_i,j}|$ represents the absolute SHAP value for the $j$-th instance of the $i$-th model. This formulation calculates the mean of the absolute SHAP values across all instances and models for the largest available sample size $N$, providing a baseline for comparison with the evolution of the explanations of individual models over time.

\subsection{Statistical Analysis}

To assess the intra-model agreement, we used Spearman correlations between sample sizes and the weighted cosine similarities of each individual model on the SHAP values computed at each step of the cross-validation. Spearman correlations are used to account for the monotonic nonlinear nature of the relationship. A similar analysis is performed for the inter-model analysis, where we computed Spearman correlations between sample sizes and weighted cosine similarities between individual models and the consensus $\bar{\phi}^{SHAP}_{consensus}$. To account for multiple comparisons, $p$-values are corrected using the Benjamini-Hochberg false discovery rate (FDR) \cite{hochbergSharperBonferroniProcedure1988}.

\section{Results}
\label{sec:results}

Having selected the top three models that maximize the Kappa coefficient on the training set, we then computed their performances on the test set. Subsequently, we compared both their internal variability (i.e., the variability caused by the cross-validation) and their overall relative variability (i.e., the variability caused by a switch of near-optimal models).

\subsection{Models performances}

\begin{figure*}[!htb]
	\centering
	\begin{tabular}{cc}
		\includegraphics[width=65mm]{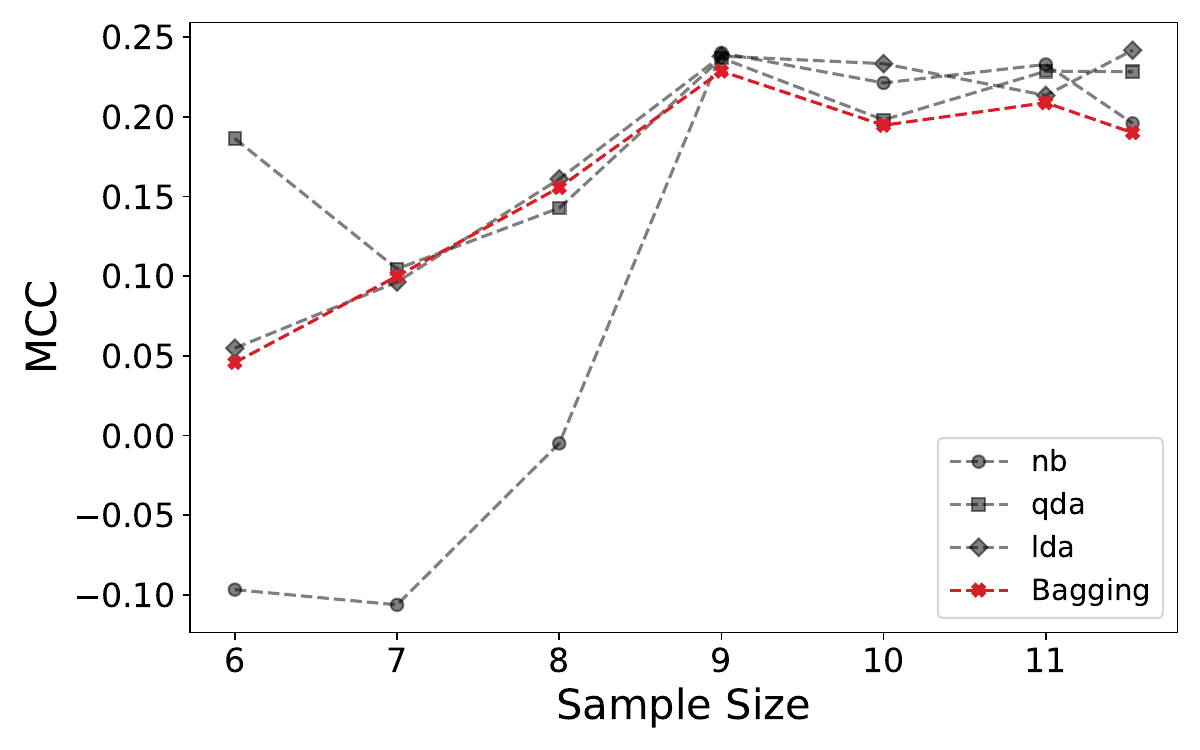} & \includegraphics[width=65mm]{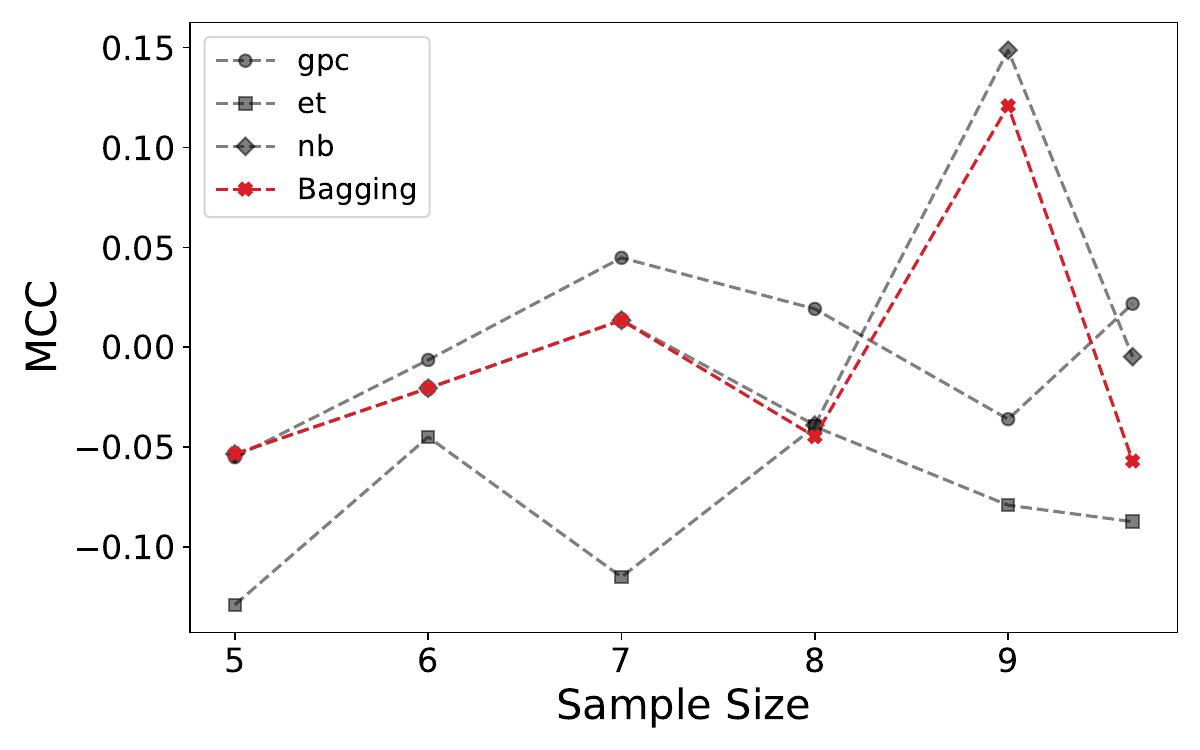} \\
		(a) Framingham dataset                                                 & (b) German dataset                                                 \\[6pt]
		\includegraphics[width=65mm]{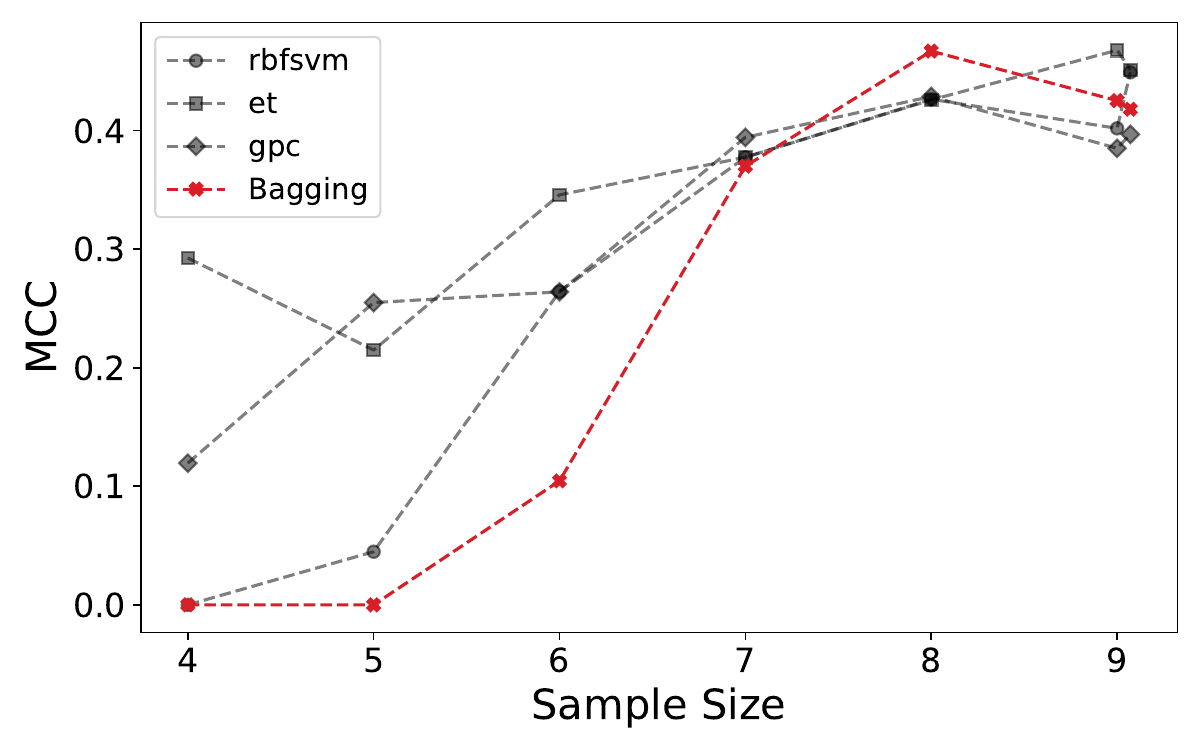}   & \includegraphics[width=65mm]{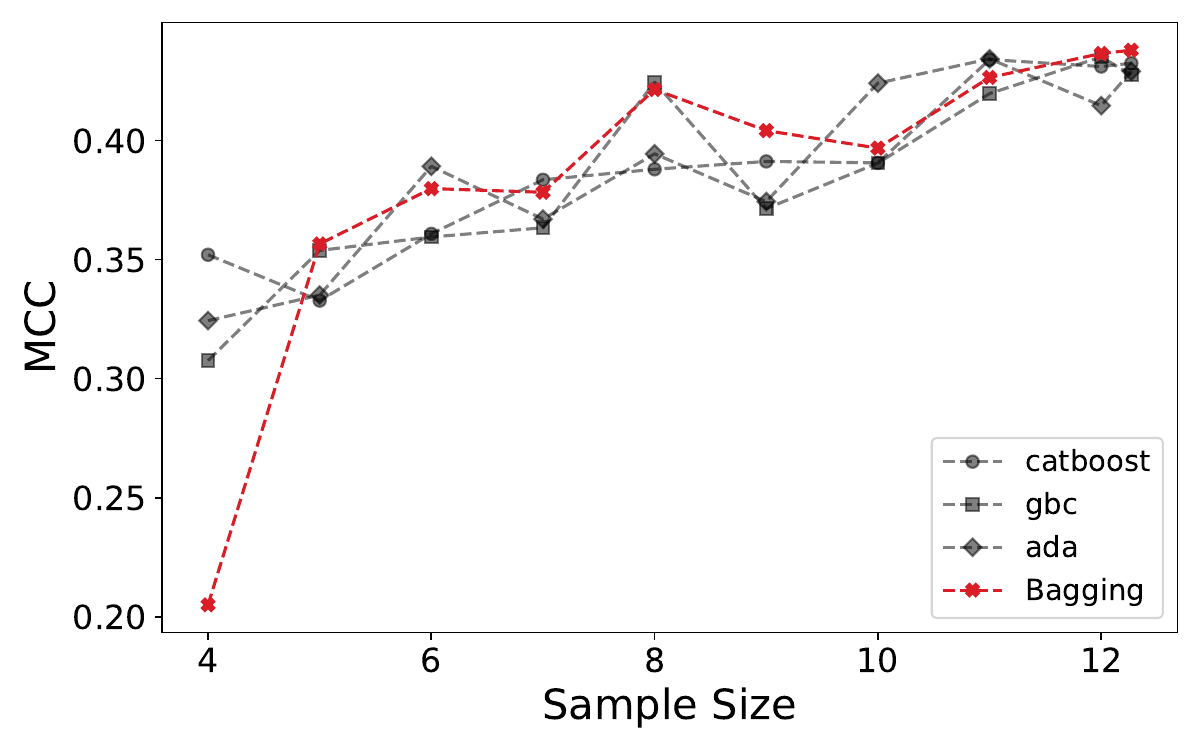} \\
		(c) Diabetes dataset                                                   & (d) Compas dataset                                                 \\[6pt]
		\multicolumn{2}{c}{\includegraphics[width=65mm]{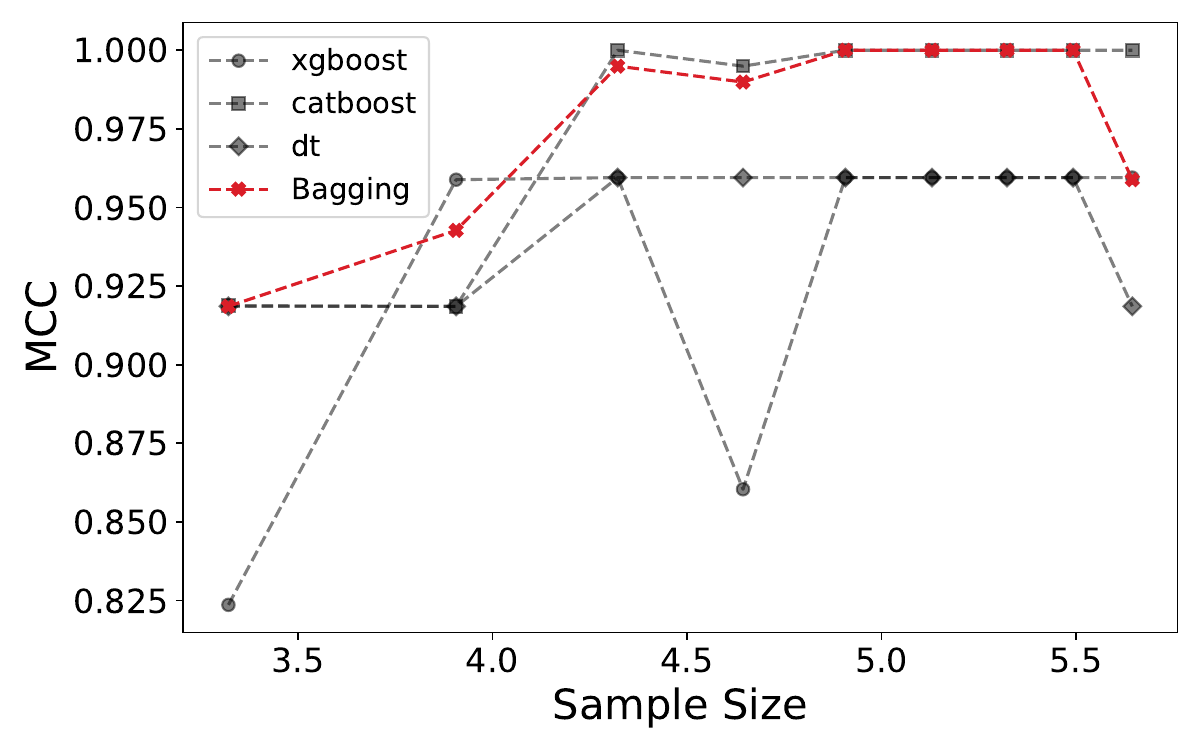} }\\
		\multicolumn{2}{c}{(e) Student dataset}
	\end{tabular}
	\caption{Impact of the sample size on model performance (Matthews Correlation Coefficient, MCC). MCC is computed on the test set during a 10-fold cross-validation. The gray lines represent individual models, and the red line represents the result when a bagging strategy is used on all models. Sample sizes a on a log2 basis.}
	\label{fig:intra_lcurve}
\end{figure*}

We implemented a 10-fold cross-validation strategy to assess the accuracy (ACC), F1, and Matthews correlation coefficients (MCC) of various models on the Framingham, German, Diabetes, Compas, and Student datasets, the results of which are illustrated in \autoref{table:perf}. The findings showed the existence of a Rashōmon set for the Framingham, Compas, Diabetes, and Student datasets, where the precision, F1 score, and MCC of the proposed models were similar. An examination of the learning curve revealed that most of the models reached a convergence state for the Framingham, Diabetes, Compas, and Student datasets, as shown in \autoref{fig:intra_lcurve}. Interestingly, it can be noted that none of the models successfully converged on the German dataset, which is characterized by high dimensionality and exhibits a pronounced class imbalance, in contrast to the Student dataset where optimal performance levels were achieved early in the learning progression, as depicted in \autoref{fig:intra_lcurve}.

\begin{table}[ht]
	\resizebox{\columnwidth}{!}{%
		\begin{tabular}{llll}
			\hline
			\textbf{}            & \textbf{ACC}          & \textbf{F1}           & \textbf{MCC}          \\ \hline
			\textit{\textbf{A.}} &                       &                       &                       \\
                baseline             & 0.663 ±0.001          & 0.231 ±0.001          & 0.200 ±0.002          \\
			nb                   & 0.833 ±0.001          & 0.268 ±0.002          & 0.196 ±0.002          \\
			qda                  & 0.828 ±0.006          & 0.318 ±0.007          & 0.228 ±0.002          \\
			lda                  & 0.808 ±0.001          & \textbf{0.354 ±0.001} & \textbf{0.242 ±0.001} \\
			\textit{bagging}     & \textbf{0.841 ±0.001} & 0.24 ±0.004           & 0.19 ±0.004           \\
			\textit{\textbf{B.}} &                       &                       &                       \\
                baseline             & 0.535 ±0.060          & 0.600 ±0.073          & -0.320 ±0.000         \\
			gpc                  & 0.595 ±0.001          & 0.714 ±0.001          & \textbf{0.022 ±0.001} \\
			et                   & \textbf{0.65 ±0.001}  & \textbf{0.783 ±0.001} & -0.072 ±0.001         \\
			nb                   & 0.5 ±0.000            & 0.59 ±0.000           & -0.005 ±0.000         \\
			\textit{bagging}     & 0.504 ±0.012          & 0.603 ±0.019          & -0.026 ±0.005         \\
			\textit{\textbf{C.}} &                       &                       &                       \\
                baseline             & 0.721 ±0.001          & 0.581 ±0.001          & 0.393 ±0.000          \\
			rbfsvm               & \textbf{0.759 ±0.003} & \textbf{0.622 ±0.018} & 0.449 ±0.009          \\
			et                   & \textbf{0.759 ±0.006} & \textbf{0.622 ±0.028} & \textbf{0.451 ±0.019} \\
			gpc                  & 0.739 ±0.001          & 0.583 ±0.0            & 0.397 ±0.001          \\
			\textit{bagging}     & 0.748 ±0.001          & 0.598 ±0.001          & 0.418 ±0.003          \\
			\textit{\textbf{D.}} &                       &                       &                       \\
                baseline             & 0.850 ±0.000          & 0.912 ±0.000          & 0.413 ±0.000          \\
			catboost             & 0.853 ±0.002          & 0.914 ±0.001          & 0.432 ±0.010          \\
			gbc                  & 0.851 ±0.005          & 0.913 ±0.003          & 0.427 ±0.019          \\
			ada                  & 0.851 ±0.003          & 0.913 ±0.002          & 0.429 ±0.007          \\
			\textit{bagging}     & \textbf{0.854 ±0.001} & \textbf{0.915 ±0.001} & \textbf{0.438 ±0.005} \\
			\textit{\textbf{E.}} &                       &                       &                       \\
                baseline             & 0.960 ±0.000          & 0.947 ±0.002          & 0.918 ±0.000          \\
			xgboost              & 0.980 ±0.001          & 0.976 ±0.001          & 0.959 ±0.001          \\
			catboost             & \textbf{1.000 ±0.000} & \textbf{1.000 ±0.000} & \textbf{1.000 ±0.000} \\
			dt                   & 0.960 ±0.001          & 0.947 ±0.001          & 0.919 ±0.001          \\
			\textit{bagging}     & 0.980 ±0.001          & 0.974 ±0.001          & 0.959 ±0.001          \\ \hline
		\end{tabular}}
        \vspace{2mm}
	\caption{Test-set performances of the sampled models on accuracy (ACC), F1, and Matthews Correlation Coefficients (MCC) on the Framingham (A), German (B), Diabetes (C), Compas (D), and Student (E) datasets. Bold results indicate best models. The baseline is a simple logistic regression.}
	\label{table:perf}
\end{table}

\subsubsection{Intra-model agreement}

\begin{figure*}[!htb]
	\centering
	\begin{tabular}{cc}
		\includegraphics[width=65mm]{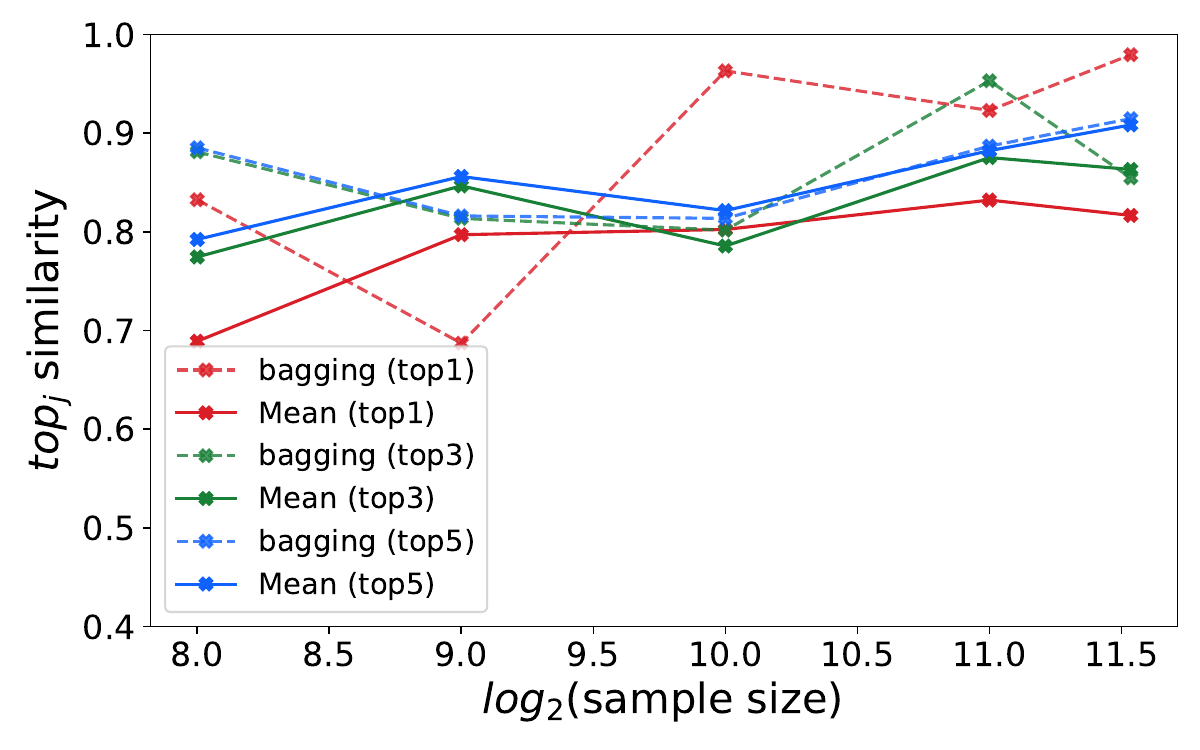} & \includegraphics[width=65mm]{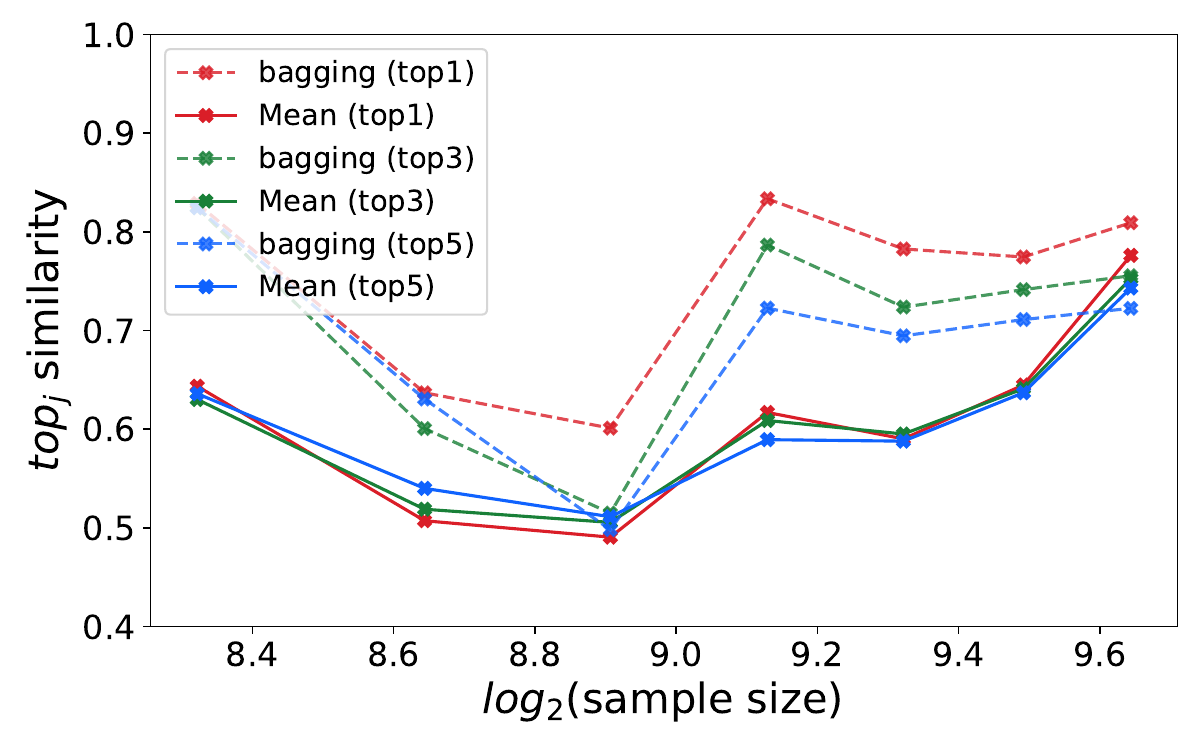} \\
		(a) Framingham dataset                                             & (b) German dataset                                             \\[6pt]
		\includegraphics[width=65mm]{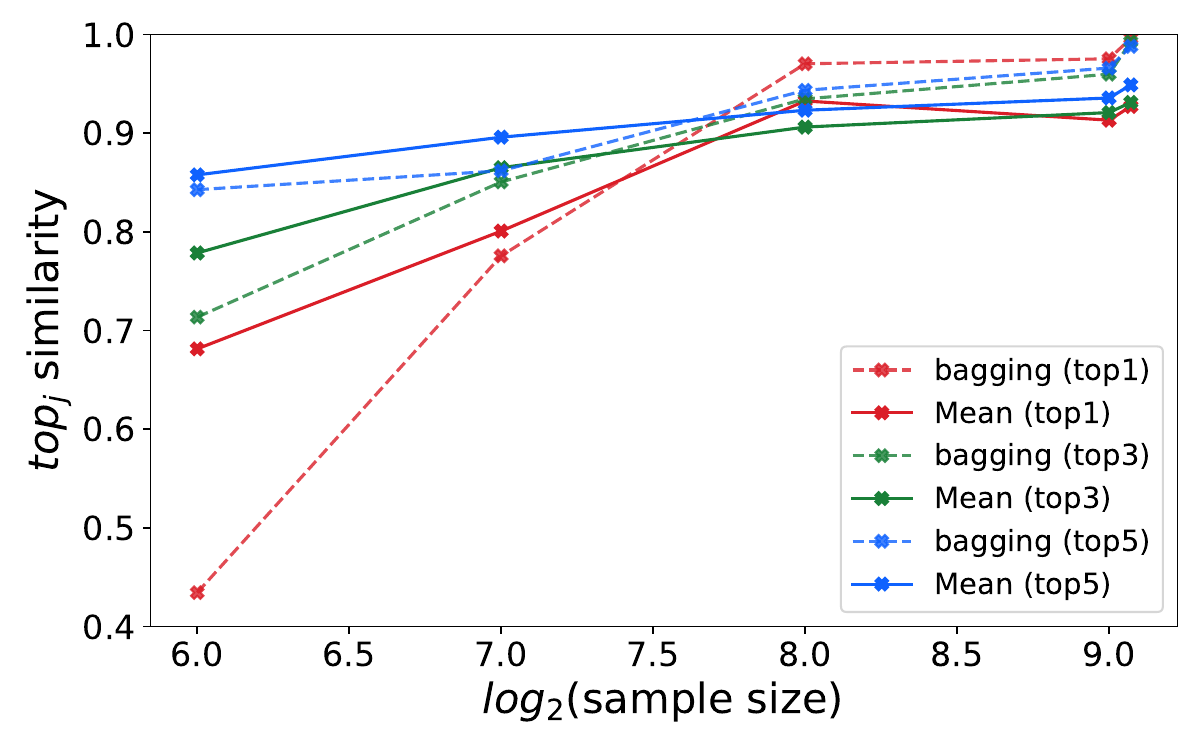}   & \includegraphics[width=65mm]{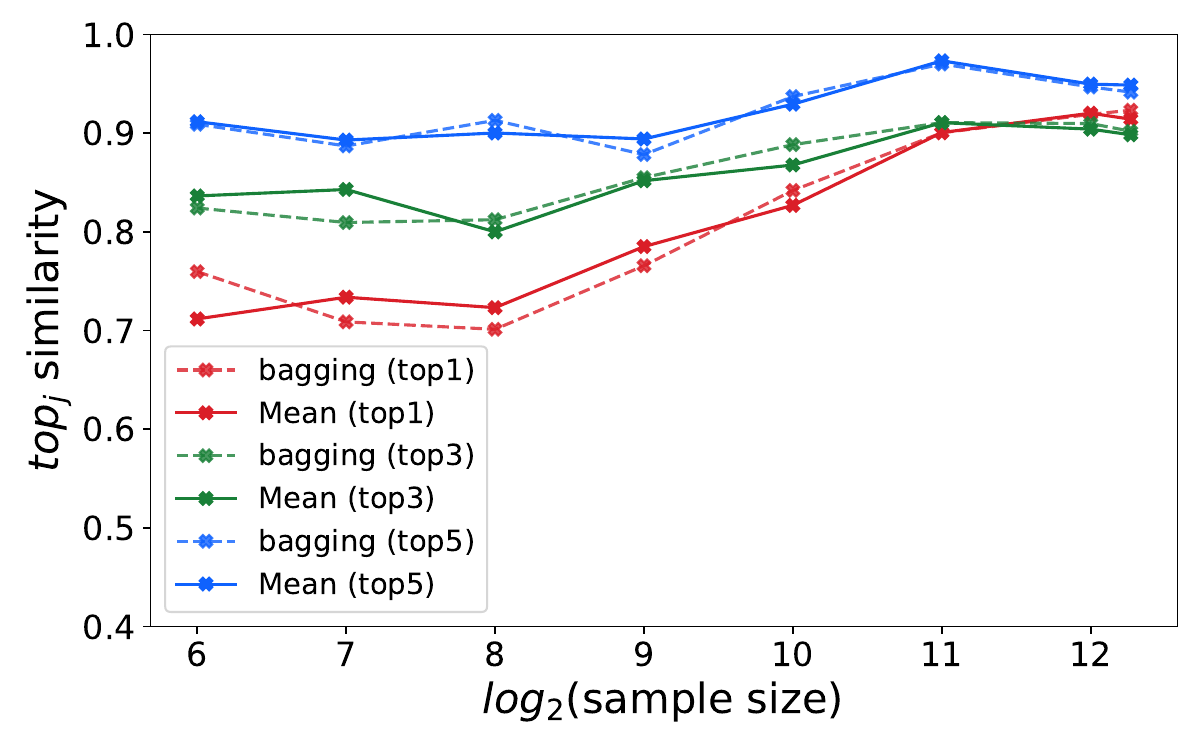} \\
		(c) Diabetes dataset                                               & (d) Compas dataset                                             \\[6pt]
		\multicolumn{2}{c}{\includegraphics[width=65mm]{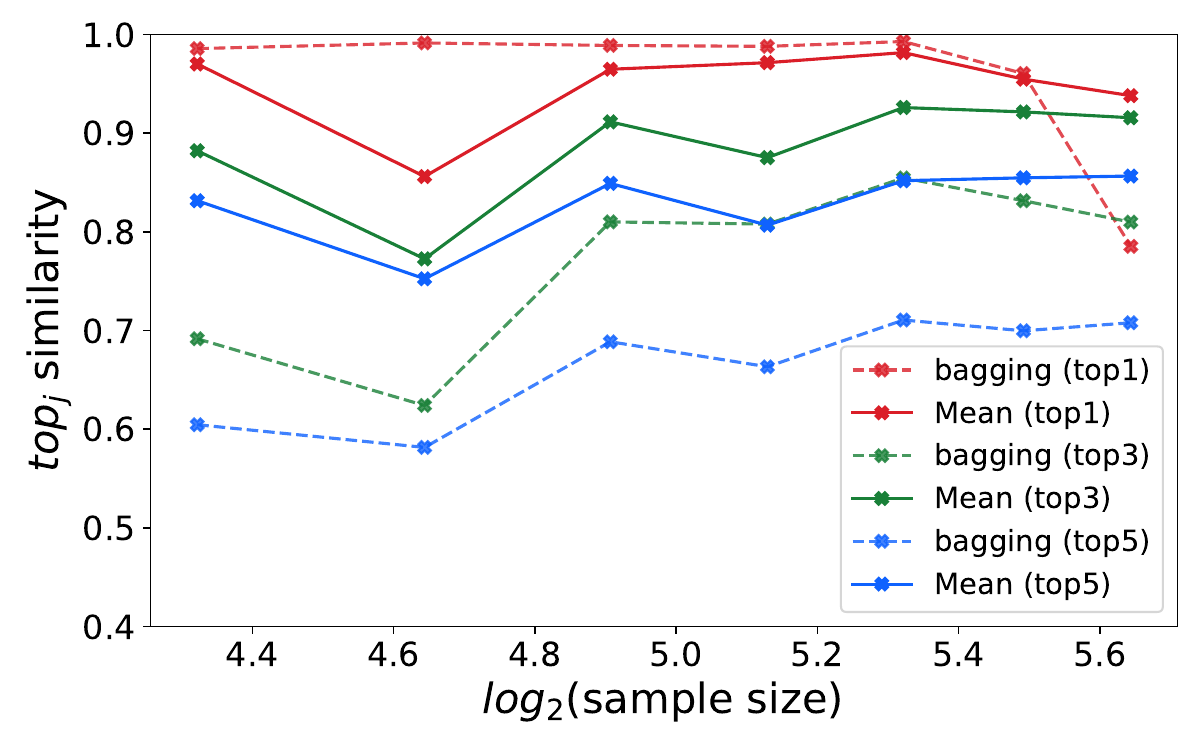} }\\
		\multicolumn{2}{c}{(e) Student dataset}
	\end{tabular}
	\caption{Impact of the sample size on $top_j$ similarity between model explanations. Explanations given for each model are computed using a 10-fold cross-validation. The dashed lines in light colors represent individual models, the solid lines represent the mean $top_j$ for $j=1$, $j=3$, and $j=5$ across all models, and their lighter counterparts represent a bagging strategy on all models.}
        \label{fig:intra_topj}
\end{figure*}

In our analyses of the Compas, Diabetes, and Framingham datasets, we found a significant correlation between sample size and intramodel agreement, as shown in \autoref{table:intra_corr}. However, this correlation was absent in the context of the Student dataset where, remarkably, even with smaller sample sizes, we recorded great levels of predictive precision and harmony, as evidenced in \autoref{table:perf}. Conversely, the German dataset exhibited persistently poor predictive accuracy throughout our experimental endeavors, a deficiency mirrored in the intra-model agreement. Overall, the $top_j$ similarity demonstrated a propensity for the $top_j$ features to converge toward uniformity, particularly for the bagging models (\autoref{fig:intra_topj}). These models displayed a higher internal agreement on the Framingham, German, and Diabetes datasets. An intriguing insight was the ability to attain a high degree of agreement with smaller samples, a level that initially receded but subsequently increased as the sample size grown.

\begin{table}[!htb]
	\centering
	\begin{tabular}{lllll}
		\hline
		           & $p$            & $p_{cor}$      & $r$   & power \\ \hline
		compas     & \textbf{0.000} & \textbf{0.000} & 0.889 & 1.000 \\
		diabetes   & \textbf{0.025} & 0.061          & 0.458 & 0.635 \\
		german     & 0.664          & 0.664          & 0.080 & 0.072 \\
		framingham & \textbf{0.045} & 0.075          & 0.412 & 0.534 \\
		student    & 0.097          & 0.122          & 0.298 & 0.389 \\ \hline
	\end{tabular}
        \vspace{2mm}
	\caption{Correlation between sample size and intra-model correlation. Results are Spearman correlations between similarities of cross-validation explanations and sample size.}
	\label{table:intra_corr}
\end{table}


\subsubsection{Inter-model agreement and consensus}

\begin{figure*}[!htb]
	\centering
	\begin{tabular}{cc}
		\includegraphics[width=65mm]{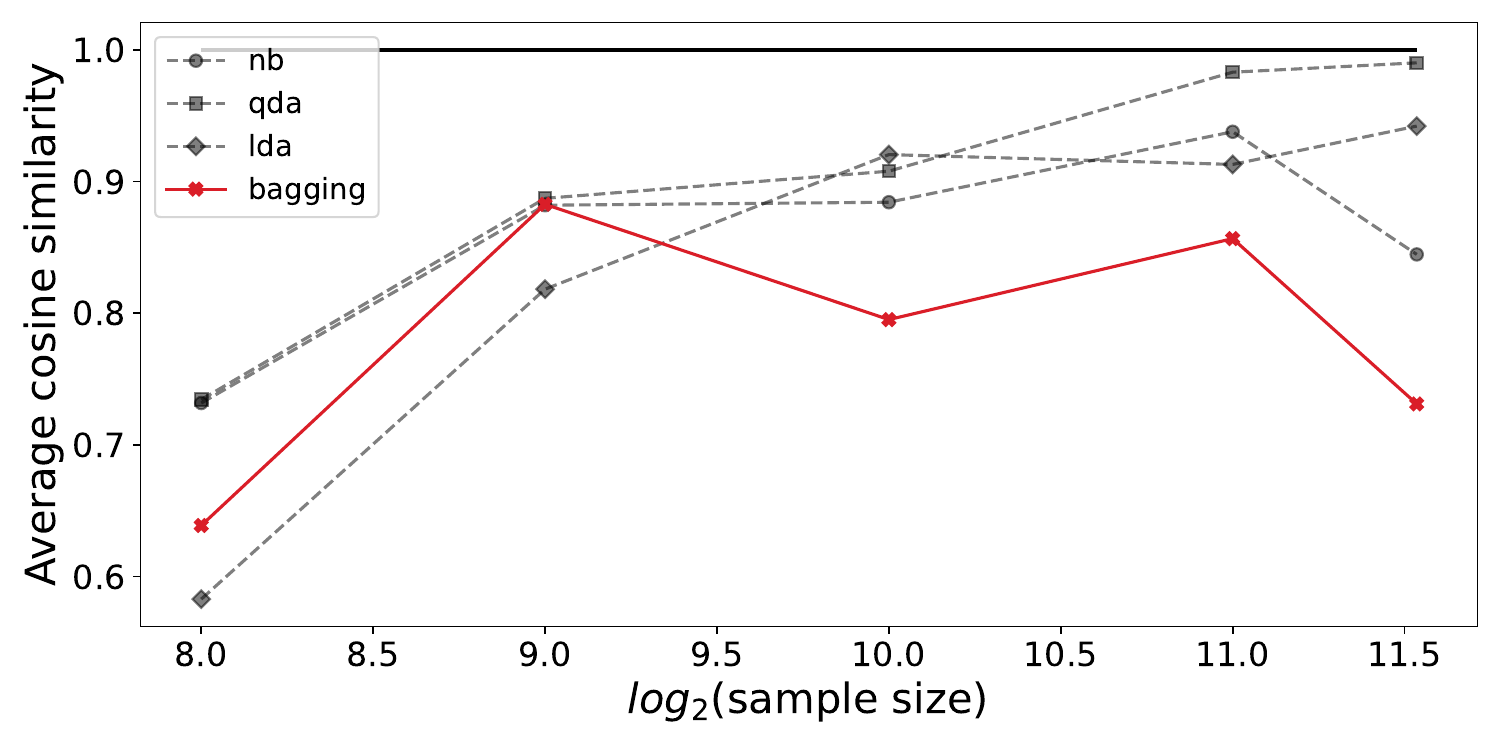} & \includegraphics[width=65mm]{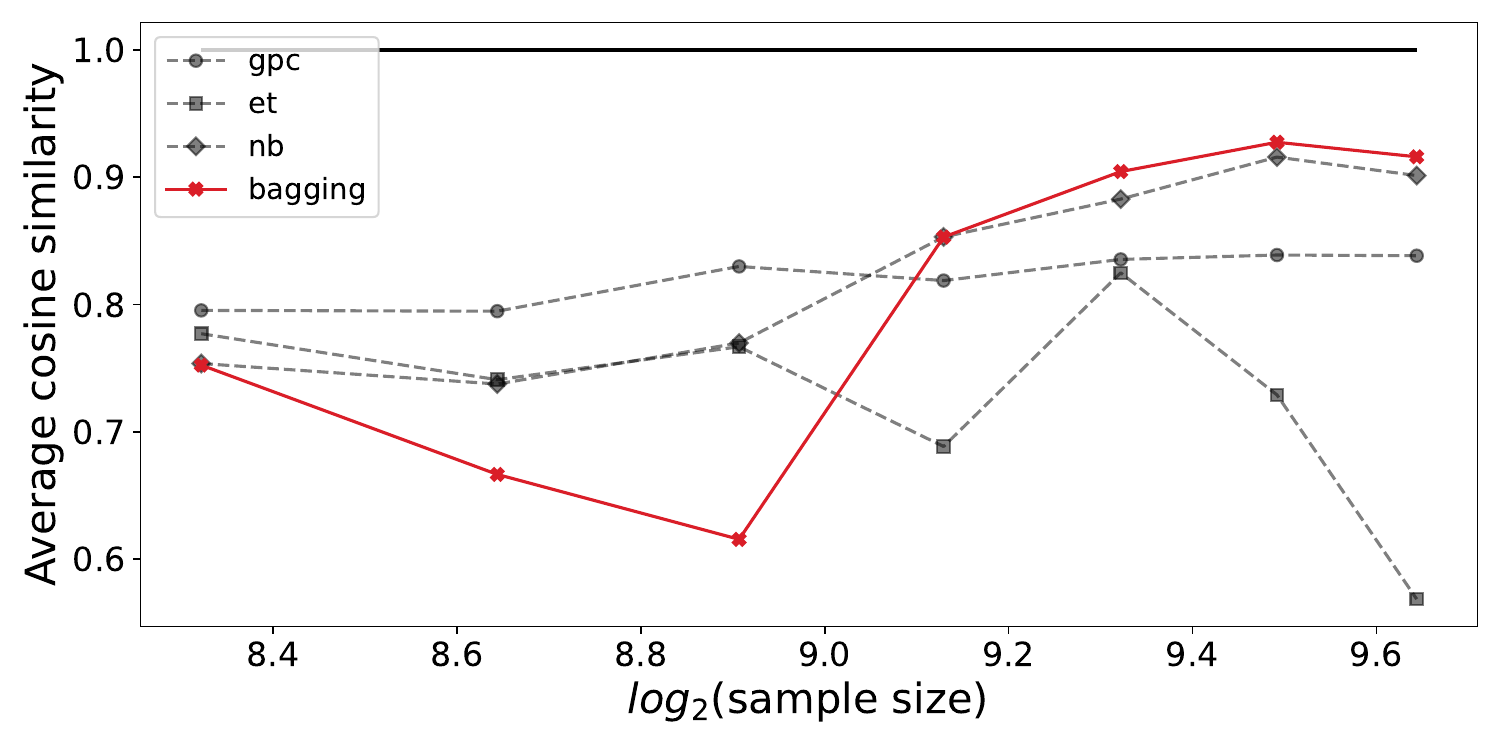} \\
		(a) Framingham dataset                                                 & (b) German dataset                                                 \\[6pt]
		\includegraphics[width=65mm]{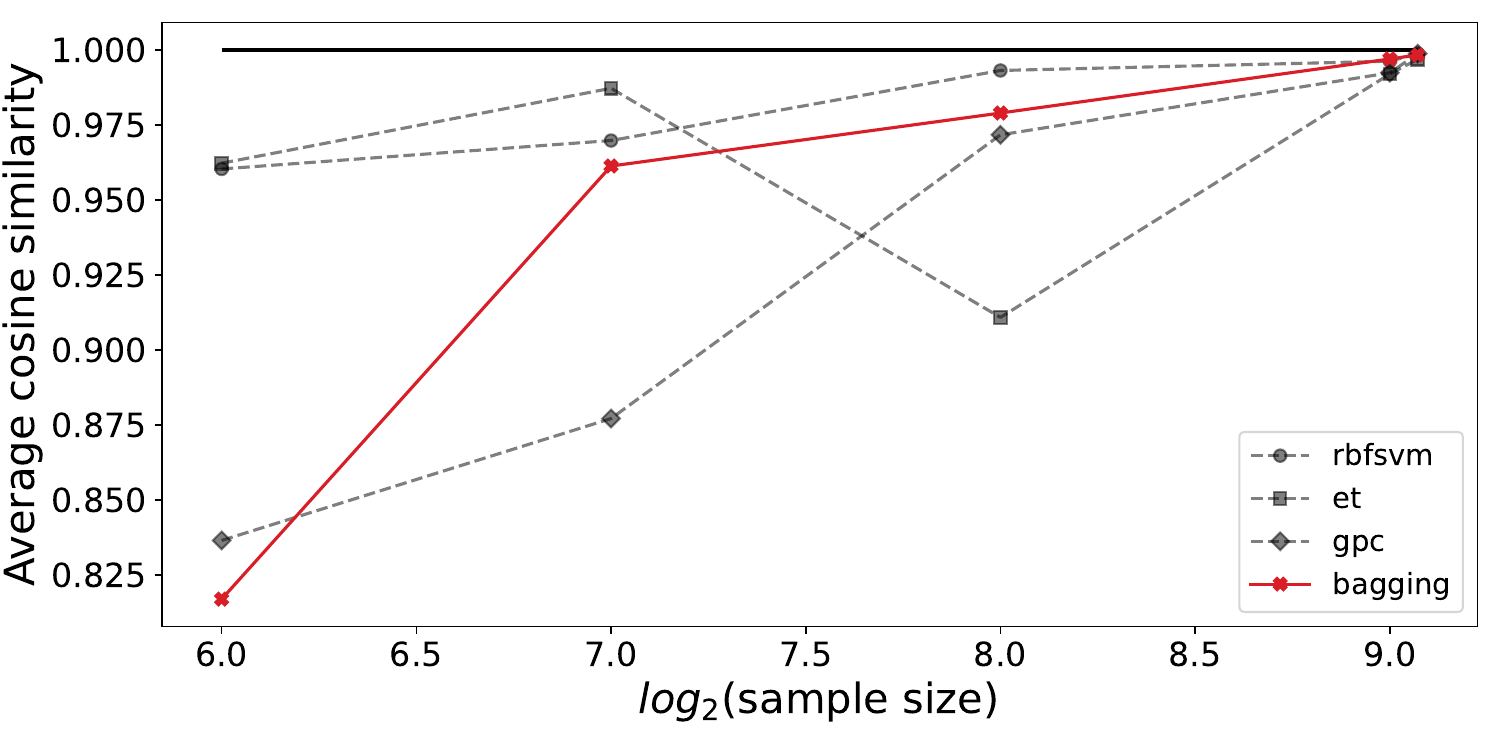}   & \includegraphics[width=65mm]{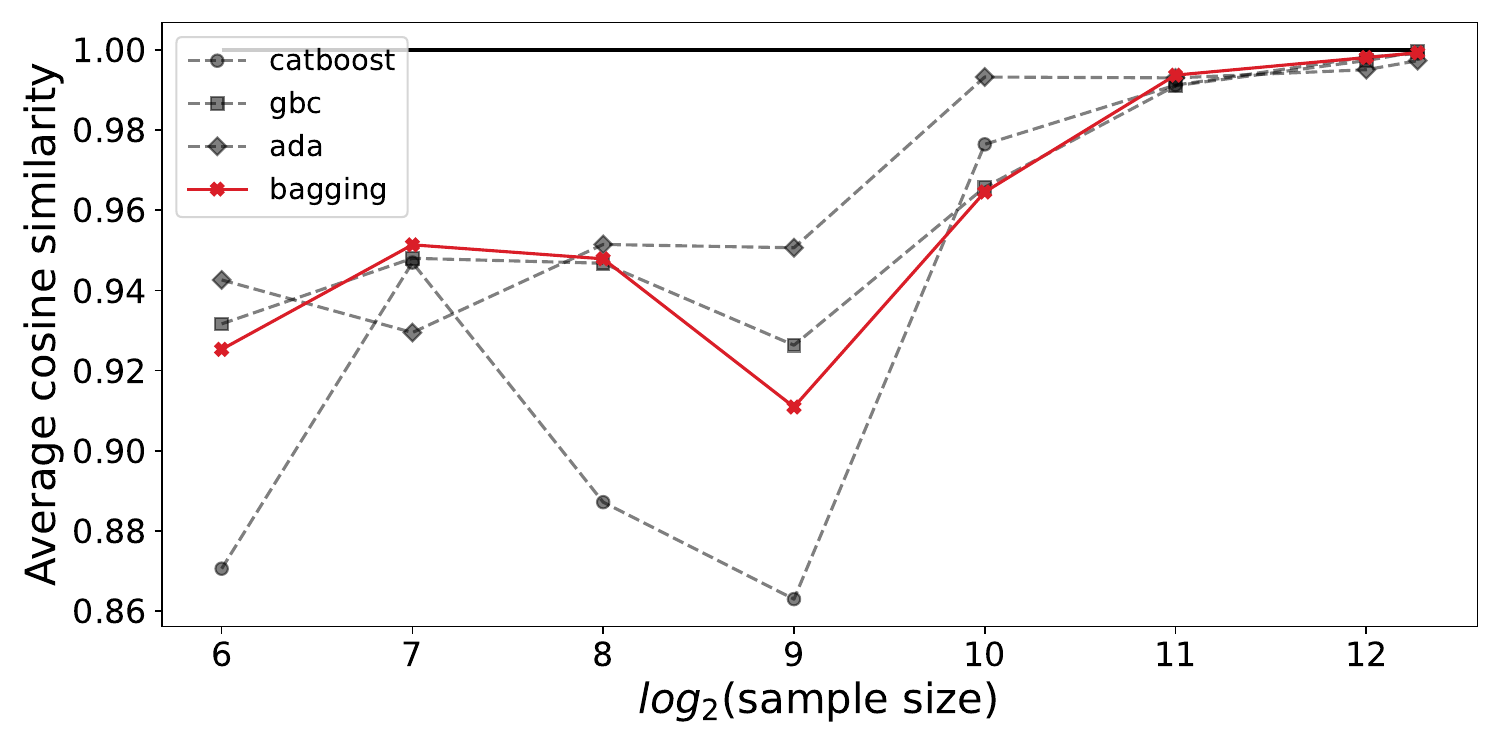} \\
		(c) Diabetes dataset                                                   & (d) Compas dataset                                                 \\[6pt]
		\multicolumn{2}{c}{\includegraphics[width=65mm]{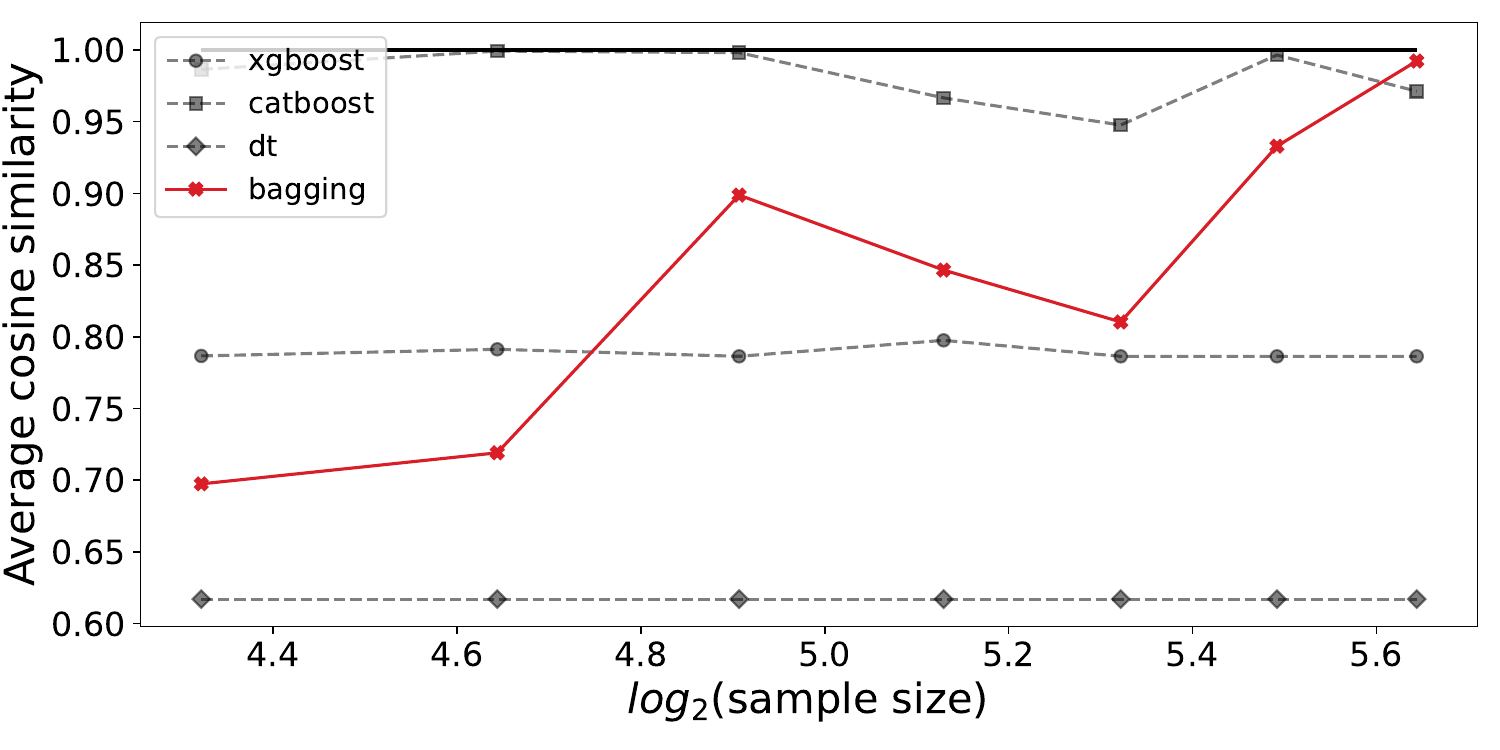} }\\
		\multicolumn{2}{c}{(e) Student dataset}
	\end{tabular}
	\caption{Impact of the convergence of models towards the best consensus available. Explanations given for each model are computed using 10-fold cross-validation. Dashed lines represent individual models; the black line represents the best consensus available (the mean of the absolute SHAP values for the biggest sample size available). The red line represents the result when using a bagging strategy on all models.}
        \label{fig:convergence}
\end{figure*}

The outcomes of the convergence towards an optimal consensus is similar to the intramodel agreement. Nearly each individual model converges towards the near-optimal consensus $\bar{\phi}^{SHAP}_{consensus}$ (\autoref{table:inter_corr} and \autoref{table:bagging_corr}), except for the student model, which demonstrates a significantly high level of similarity even for small sample sizes. As for the German model, it follows a U-shaped pattern with respect to the sample size (\autoref{fig:convergence}).

\begin{table}[!htb]
	\centering
	\begin{tabular}{lllll}
		\hline
		           & $p$            & $p_{cor}$      & $r$    & power \\ \hline
		compas     & \textbf{0.000} & \textbf{0.000} & 0.766  & 0.999 \\
		diabetes   & \textbf{0.000} & \textbf{0.000} & 0.862  & 0.999 \\
		german     & 0.111          & 0.139          & 0.334  & 0.366 \\
		framingham & \textbf{0.000} & \textbf{0.000} & 0.875  & 1.000 \\
		student    & 0.654          & 0.654          & -0.096 & 0.073 \\ \hline
	\end{tabular}
        \vspace{2mm}
	\caption{Correlation between sample size and inter-model correlation convergence towards the best consensus. Results are Spearman correlations between similarities of all models and $\bar{\phi}^{SHAP}_{consensus}$ and sample size.}
        \label{table:inter_corr}
\end{table}

\begin{table}[!htb]
	\centering
	\begin{tabular}{lllll}
		\hline
		           & $p$            & $p_{cor}$      & $r$   & power \\ \hline
		compas     & 0.067          & 0.084          & 0.633 & 0.483 \\
		diabetes   & \textbf{0.000} & \textbf{0.000} & 1.000 & 1.000 \\
		german     & \textbf{0.037} & 0.061          & 0.738 & 0.604 \\
		framingham & 0.329          & 0.329          & 0.486 & 0.171 \\
		student    & \textbf{0.002} & \textbf{0.005} & 0.905 & 0.936 \\ \hline
	\end{tabular}
        \vspace{2mm}
	\caption{Correlation between sample size and bootstrap aggregation convergence towards the best consensus. Results are Spearman correlations between similarities of a bagging strategy of all $top_3$ models and $\bar{\phi}^{SHAP}_{consensus}$ and sample size.}
        \label{table:bagging_corr}
\end{table}

\section{Discussion}
\label{sec:discussion}


The problem of the existence of multiple high-performing models relying on different features --- the Rashōmon effect --- poses significant challenges to derive reliable knowledge from machine learning systems. In this study, our aim was to examine the influence of sample size on models within a Rashōmon set using a model-agnostic explainability technique. Our research revolved around two main aspects: (i) the enhancement of explainability in relation to sample size, and (ii) the convergence of explanations from diverse models in the Rashōmon set, leading to a consensus when the sample size reaches a sufficient size. Additionally, we explored the characteristics of bagging strategies within the same scenarios to gain additional information. The key findings of this study demonstrate the substantial impact of sample size on the reliability and agreement of explanations from machine learning models that exhibit a Rashōmon effect (\autoref{table:intra_corr}, \autoref{table:inter_corr}, and \autoref{table:bagging_corr}). Our experiments in five public data sets revealed that explanations derived from subsets with fewer than 128 samples showed high variability in cross-validation, indicating spuriousness (\autoref{fig:intra_topj} and \autoref{fig:convergence}). This effectively limits the actionable knowledge that can be extracted from any single model's interpretations. However, as the sample size increased, the variance in similarity diminished and models converged towards a unified explanation.


Our first experiment was primarily devoted to the study of model robustness when the training set is perturbed through cross-validation, a property we named internal agreement. All models, except the one used for the German dataset, demonstrated convergence towards an acceptable level of classification accuracy (\autoref{fig:intra_lcurve} and \autoref{table:perf}), while manifesting a strong correlation between sample size and weighted cosine similarity (\autoref{table:intra_corr}). However, it is important to highlight two exceptions. The first pertains to the methodologies employed on the German dataset, which were unable to identify suitable models within the Rashōmon set, implying that no models were successful in determining a valid solution, resulting in a lack of convergence in internal agreement. The second exception is linked to the Student dataset, where convergence was observed remarkably early in the learning trajectory (\autoref{fig:intra_lcurve}e), indicating that a high degree of internal agreement was achieved at the beginning of the experiment itself. An interesting insight was the nonlinear association between the sample size and the intra-model consensus. At smallS sample sizes, high similarity could be achieved (\autoref{fig:intra_topj}b) despite a poor predictive capacity (\autoref{fig:intra_lcurve}b), which subsequently experienced a decline, only to ascend once again with the addition of more data. This aligns with previous research indicating false confidence in underspecified models \cite{waldCalibrationOutofdomainGeneralization2022}, drawing a parallel to the Dunning-Kruger effect observed in human learning. Although the agreement on the German dataset seems to increase, this suggests that more data is needed to perform our analysis on our dataset. The second important point that this experiment highlights is the regular superiority of bagging approaches, where they showed higher internal agreement at large sample sizes for 3 of our 5 datasets (\autoref{fig:intra_topj}), but this was not always the case at low sample sizes, especially for the Diabetes dataset. This first experiment allows us to say that while sample size has a positive impact on a model's internal agreement, performance is not always correlated to agreement in explanations, highlighting the need to perform additional analysis before the explanation of machine learning models. Moreover, contrary to popular belief, bagging is not always the best strategy --- especially at low sample sizes --- regarding explainability.


The supplementary experiment engaged in an in-depth exploration concerning the coherence among various models and their convergence towards a near-optimal consensus, --- the average explanation of high-performing models at the highest sample size. Once again, we observed a positive correlation between the sample size and the agreement among the high-performance models (\autoref{table:inter_corr} and \autoref{table:bagging_corr}). Analogous to the intra-model agreement, the experiment failed to exhibit any correlation within the Student dataset due to the previously delineated complications. The convergence of inter-model explanations reaffirms the hypothesis that large datasets attenuate the Rashōmon effect and facilitate reliable knowledge extraction \cite{del_giudice_prediction-explanation_2021}. Despite the fact that the bagging ensembles did not always converged better than the individual models (\autoref{fig:convergence}a, c, and d), they converged towards the optimal solution, while some models remained stuck within their respective basins (\autoref{fig:convergence}e). This secondary experiment allows the conclusion that the learning process functions herein as a refinement of feature importance, a process whereby an increase in sample size triggers a convergence towards a singular, coherent explanation of the predictions. However, it is interesting to observe the phenomenon that underscores the disparities between a consensus derived from explanations coming from diverse models, and the explanation of a bagging procedure over the exact same models (e.g. \autoref{fig:convergence}a). Although it was generally not possible to differentiate between the predictive performance of individual models and the bagging procedure (\autoref{fig:intra_lcurve}), this research does not offer the means to draw a definitive conclusion about which is superior: should we favor the explanation derived from an aggregation of multiple models, or a consensus predicated upon the explanations of the individual models? However, the conclusion we can draw is that both were similar for 4 datasets in the 5 we studied, meaning that our method should be appropriate in both cases.

However, there are some limitations to the generalization of these findings. First, this study limited itself to certain types of models, such as linear models or random forests. Testing other complex models, such as neural networks, could reveal different behaviors and challenges. Extending this research to neural network models presents challenges due to their complexity, but also opportunities to mitigate the Rashōmon effect through transfer learning. As shown by \cite{neyshaburWhatBeingTransferred2021}, different initializations of the same architecture can learn distinct solutions. Transfer learning provides more consistent representations and could align explanations between models, because the learning on the specific task starts in an already nonrandom state. Questions such as \textit{``does the surrogate task impact explanations?''} or \textit{``are explanations found through transfer learning better than those of models trained from scratch?''} remain to be answered. Testing the sample size effects shown here in deep learning scenarios with and without transfer learning would be an impactful extension.

Second, while the datasets covered various sizes and domains, more extensive experiments are needed on diverse real-world problems, which could include regression problems. Regarding those datasets, the Student task was definitely too simple for the correct assessment of our methodology, even if it allowed us to draw interesting conclusions. However, the German dataset clearly requested a modified methodology, certainly a dimensionality reduction, such as PCA, or a class-imbalance correction, such as SMOTE. We intentionally chose not to adopt this modified methodology to stay consistent with the other analysis. Also, PCA would have complexified our SHAP workflow as we would have had to interpret its principal components. However, this raises a relevant link between the curse of dimensionality and overconfidence issues in machine learning models as described in other works (e.g. \cite{catalbasInvestigationRelationshipCurse2020}).

Finally, while this study used SHAP, a model-agnostic approach, further research could explore how model-specific explanation methods such as attention maps or Grad-CAM behave in the context of the Rashōmon effect. As model-specific techniques provide insights into a model's internal representations, they may reveal divergences between architectures that model-agnostic methods cannot access. Comparing the convergence patterns of both approaches could offer additional validation and confidence in the explanations. Furthermore, the evaluation of other post-hoc explanation techniques such as LIME could reveal different insights. For example, LIME perturbs inputs and trains interpretable local models, which may show higher variability at low sample sizes. Testing multiple perturbation and explanation strategies could indicate which are the most robust for reliable explanations from underspecified models.

Despite these constraints, this work has meaningful practical implications. The results suggest that sample sizes below 100 can lead to unreliable explanations that lack consensus between equally performing models. We argue that our methodology provides a kind of power analysis to determine sufficient data to trust explanations. Furthermore, the variability at low sizes indicates that conclusions drawn from small datasets could be spurious and should be validated. In general, these findings can guide machine learning practitioners in selecting appropriate data volumes, models, and explanation techniques for their applications.

Future work should explore ensembles such as bagging for explanation robustness across broader model types, data domains, and explanation methods. Testing the Rashōmon effect in online learning settings where models are continuously retrained would also have a great impact. As interpretations become increasingly critical for trustworthy AI, developing a rigorous understanding of how to evaluate, improve, and trust model explanations remains an essential challenge. The approaches explored here --- leveraging sample size, ensembles, consensus-finding, and variability quantification --- point towards principled pathways for eliciting knowledge from ambiguous models. This study provides an initial data-driven perspective on this important problem.

\section{Conclusion}
\label{sec:conclusion}

This study examined how sample size impacts the explainability of models that exhibit Rashōmon effect. Experiments on multiple datasets revealed explanations become more consistent as sample size grows, with variability limiting reliability below 100 samples. Key takeaways for practitioners include: 1) larger data volumes attenuate the Rashōmon effect and improve explanation consensus, 2) explanations derived from limited data may be spurious and require validation, 3) bagging ensembles can enhance agreement between models. Overall, these findings provide guidance on selecting appropriate sample sizes, models, and explanation techniques when interpreting machine learning systems to ensure credible and actionable knowledge.



\normalsize
\bibliography{references}


\end{document}